# Deep EHR: A Survey of Recent Advances in Deep Learning Techniques for Electronic Health Record (EHR) Analysis

Benjamin Shickel[1], Patrick J. Tighe[2], Azra Bihorac[3], and Parisa Rashidi[4]

*Abstract*—The past decade has seen an explosion in the amount of digital information stored in electronic health records (EHR). While primarily designed for archiving patient information and performing administrative healthcare tasks like billing, many researchers have found secondary use of these records for various clinical informatics applications. Over the same period, the machine learning community has seen widespread advances in the field of deep learning. In this review, we survey the current research on applying deep learning to clinical tasks based on EHR data, where we find a variety of deep learning techniques and frameworks being applied to several types of clinical applications including information extraction, representation learning, outcome prediction, phenotyping, and de-identification. We identify several limitations of current research involving topics such as model interpretability, data heterogeneity, and lack of universal benchmarks. We conclude by summarizing the state of the field and identifying avenues of future deep EHR research.

## I. INTRODUCTION

OVER the past 10 years, hospital adoption of electronic health record (EHR) systems has skyrocketed, in part due to the Health Information Technology for Economic and Clinical Health (HITECH) Act of 2009, which provided $30 billion in incentives for hospitals and physician practices to adopt EHR systems [1]. According to the latest report from the Office of the National Coordinator for Health Information Technology (ONC), nearly 84% of hospitals have adopted at least a basic EHR system, a 9-fold increase since 2008 [2]. Additionally, office-based physician adoption of basic and certified EHRs has more than doubled from 42% to 87% [3].

EHR systems store data associated with each patient encounter, including demographic information, diagnoses, laboratory tests and results, prescriptions, radiological images, clinical notes, and more [1]. While primarily designed for improving healthcare efficiency from an operational standpoint, many studies have found secondary use for clinical informatics applications [4], [5]. In particular, the patient data contained in EHR systems has been used for such tasks as medical concept extraction [6], [7], patient trajectory modeling [8], disease inference [9], [10], clinical decision support systems [11], and more (Table I).

[1]B. Shickel is with the Department of Computer & Information Science, University of Florida, Gainesville, Florida, USA. shickelb@ufl.edu
[2]P. Tighe MD MS is with the Department of Anesthesiology, College of Medicine, University of Florida, Gainesville, Florida, USA. ptighe@anest.ufl.edu
[3]A. Bihorac MD MS is with the Department of Nephrology, College of Medicine, University of Florida, Gainesville, Florida, USA. Azra.Bihorac@medicine.ufl.edu
[4]P. Rashidi is with the J. Crayton Pruitt Department of Biomedical Engineering, University of Florida, Gainesville, Florida, USA. parisa.rashidi@ufl.edu

TABLE I
SEVERAL RECENT DEEP EHR PROJECTS.

| Project | Deep EHR Task | Ref. |
| --- | --- | --- |
| DeepPatient | Multi-outcome Prediction | Miotto [14] |
| Deepr | Hospital Re-admission Prediction | Nguyen [19] |
| DeepCare | EHR Concept Representation | Pham [20] |
| Doctor AI | Heart Failure Prediction | Choi [21] |
| Med2Vec | EHR Concept Representation | Choi [22] |
| eNRBM | Suicide risk stratification | Tran [23] |

Until the last few years, most of the techniques for analyzing rich EHR data were based on traditional machine learning and statistical techniques such as logistic regression, support vector machines (SVM), and random forests [12]. Recently, deep learning techniques have achieved great success in many domains through deep hierarchical feature construction and capturing long-range dependencies in data in an effective manner [13]. Given the rise in popularity of deep learning approaches and the increasingly vast amount of patient data, there has also been an increase in the number of publications applying deep learning to EHR data for clinical informatics tasks [14]–[18] which yield better performance than traditional methods and require less time-consuming preprocessing and feature engineering.

In this paper, we review the specific deep learning techniques employed for EHR data analysis and inference, and discuss the concrete clinical applications enabled by these advances. Unlike other recent surveys [24] which review deep learning in the broad context of health informatics applications ranging from genomic analysis to biomedical image analysis, our survey is focused exclusively on deep learning techniques tailored to EHR data. Contrary to the selection of singular, distinct applications found in these surveys, EHR-based problem settings are characterized by the heterogeneity and structure of their data sources (Section II) and by the variety of their applications (Section V).

### A. Search strategy and selection criteria

We searched Google Scholar for studies published up to and including August 2017. All searches included the term "electronic health records" or "electronic medical records" or "EHR" or "EMR", in conjunction with either "deep learning" or the name of a specific deep learning technique (Section IV). Figure 1 shows the distribution of the number of publications per year in a variety of areas relating to deep EHR. The top subplot of Figure 1 contains a distribution of studies for the search "deep learning" "electronic health records", which highlights the overall yearly increase in the volume of publications relating to deep learning and EHR. The final two subplots contain the same search in conjunction with additional terms relating to either applications (center) or techniques (bottom). For these searches, we include variations of added terms as an OR clause, for example: "recurrent neural network" OR "RNN" "deep learning" "electronic health records". As the overall volume of publications is relatively low given the recency of this field, we manually reviewed all articles and included the most salient and archetypal deep EHR publications in the remainder of this survey.

We begin by reviewing EHR systems in Section II. We then explain key machine learning concepts in Section III, followed by deep learning frameworks in Section IV. Next, we look at recent applications of deep learning for EHR data analysis in Section V. Finally, we conclude the paper by identifying current challenges and future opportunities in Section VII.

## II. ELECTRONIC HEALTH RECORD SYSTEMS (EHR)

The use of EHR systems has greatly increased in both hospital and ambulatory care settings [2], [3]. EHR usage at hospitals and clinics has the potential to improve patient care by minimizing errors, increasing efficiency, and improving care coordination, while also providing a rich source of data for researchers [25]. EHR systems can vary in terms of functionality, and are typically categorized into basic EHR without clinical notes, basic EHR with clinical notes, and comprehensive systems [2]. While lacking more advanced functionality, even basic EHR systems can provide a wealth of information on patient's medical history, complications, and medication usage.

Since EHR was primarily designed for internal hospital administrative tasks, several classification schema and controlled vocabularies exist for recording relevant medical information and events. Some examples include diagnosis codes such as the International Statistical Classification of Diseases and Related Health Problems (ICD), procedure codes such as the Current Procedural Terminology (CPT), laboratory observations such as the Logical Observation Identifiers Names and Codes (LOINC), and medication codes such as RxNorm. Several examples are shown in Table II. These codes can vary between institutions, with partial mappings maintained by resources such as the United Medical Language System (UMLS) and the Systemized Nomenclature of Medicine - Clinical Terms (SNOMED CT). Given the large array of schemata, harmonizing and analyzing data across terminologies and between institutions is an ongoing area of research. Several of the deep EHR systems in this paper propose forms of clinical code representation that lend themselves more easily to cross-institution analysis and applications.

EHR systems store several types of patient information, including demographics, diagnoses, physical exams, sensor measurements, laboratory test results, prescribed or administered medications, and clinical notes. One of the challenges in working with EHR data is the heterogeneous nature by which it is represented, with data types including: **(1)** *numerical quantities* such as body mass index, **(2)** *datetime objects* such as date of birth or time of admission, **(3)** *categorical values* such as ethnicity or codes from controlled vocabularies like ICD-10 (formerly ICD-9) diagnoses or CPT procedures, and **(4)** *natural language free-text* such as progress notes or discharge summaries. Additionally, these data types can be ordered chronologically to form the basis for **(5)** *derived time series* such as perioperative vital sign signals or multimodal patient history. While other biomedical data such as medical images or genomic information exist and are covered in recent relevant articles [24], [26], [27], in this survey we focus on these five data types found in most modern EHR systems.

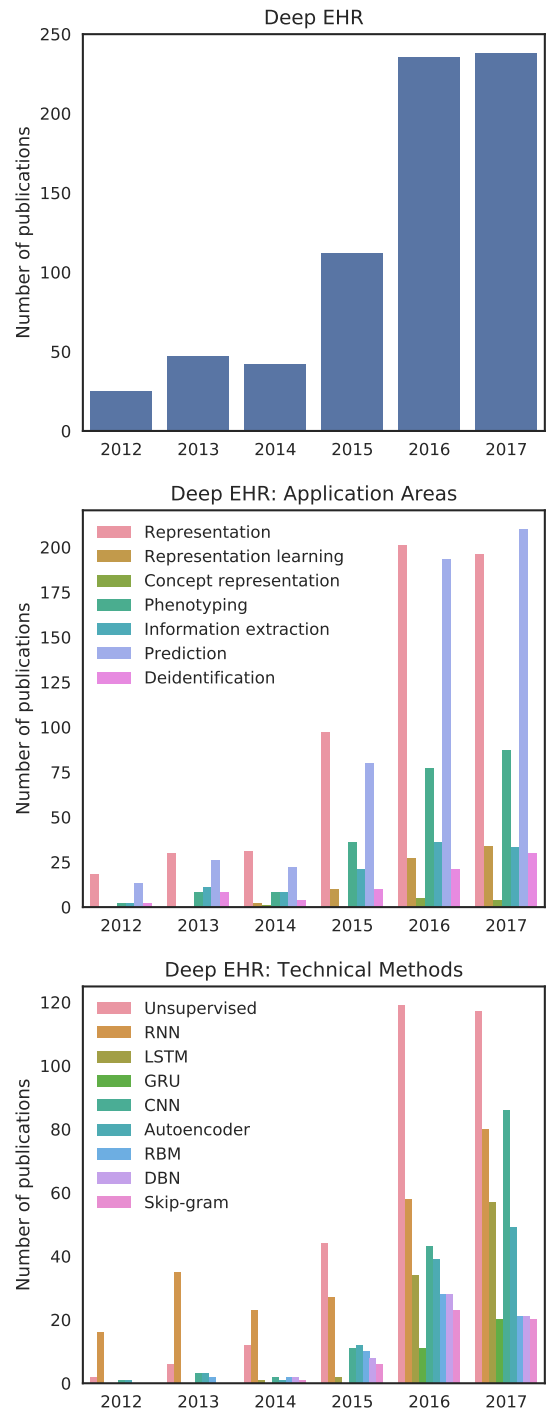

Fig. 1. Trends in the number of Google Scholar publications relating to deep EHR through August 2017. The top distribution shows overall results for "deep learning" and "electronic health records". The bottom two distributions show these same terms in conjunction with a variety of specific application areas and technical methods. Large yearly jumps are seen for most terms beginning in 2015.

TABLE II
EXAMPLE CLASSIFICATION SCHEMA FOR DIAGNOSES, PROCEDURES, LABORATORY TESTS, AND MEDICATIONS.

| Schema | Number of Codes | Examples |
| --- | --- | --- |
| ICD-10 *(Diagnosis)* | 68,000 | - J9600: Acute respiratory failure<br>- I509: Heart failure<br>- I5020: Systolic heart failure |
| CPT *(Procedures)* | 9,641 | - 72146: MRI Thoracic Spine<br>- 67810: Eyelid skin biopsy<br>- 19301: Partial mastectomy |
| LOINC *(Laboratory)* | 80,868 | - 4024-6: Salicylate, Serum<br>- 56478-1: Ethanol, Blood<br>- 3414-0: Buprenorphine Screen |
| RxNorm *(Medications)* | 116,075 | - 161: Acetaminophen<br>- 7052: Morphine<br>- 1819: Buprenorphine |

## III. MACHINE LEARNING OVERVIEW

Machine learning approaches can be broadly divided into two major categories: *supervised* and *unsupervised learning*. Supervised learning techniques involve inferring a mapping function $y = f(x)$ from inputs $x$ to outputs $y$. Examples of supervised learning tasks include regression and classification, with algorithms including logistic regression and support vector machines. In contrast, the goal of unsupervised machine learning techniques is to learn interesting properties about the distribution of $x$ itself. Examples of unsupervised learning tasks include clustering and density estimation.

The *representation* of inputs is a fundamental issue spanning all types of machine learning frameworks. For each data point, sets of attributes known as *features* are extracted to be used as input to machine learning techniques. In traditional machine learning, these features are hand-crafted based on domain knowledge. One of the core principles of deep learning is the automatic data-oriented feature extraction, as discussed in the following subsection.

## IV. DEEP LEARNING OVERVIEW

Deep learning encompasses a wide variety of techniques. In this section, we provide a brief overview of the most common deep learning approaches. For each specific architecture, we highlight a key equation that illustrates its fundamental method of operation. For a more detailed overview, please refer to the comprehensive work of Goodfellow et al. [28].

The most central idea in deep learning is that of *representation*. Traditionally, input features to a machine learning algorithm must be hand-crafted from raw data, relying on practitioner expertise and domain knowledge to determine explicit patterns of prior interest. The engineering process of creating, analyzing, selecting, and evaluating appropriate features can be laborious and time consuming, and is often thought of as a "black art" [29] requiring creativity, trial-and-error, and oftentimes luck. In contrast, deep learning techniques learn optimal features directly from the data itself, without any human guidance, allowing for the automatic discovery of latent data relationships that might otherwise be unknown or hidden.

Complex data representation in deep learning is often expressed as compositions of other, simpler representations. For instance, recognizing a human in an image can involve finding representation of edges from pixels, contours and corners from edges, and facial features from corners and contours [28]. This notion of unsupervised hierarchical representation of increasing complexity is a recurring deep learning theme.

The vast majority of deep learning algorithms and architectures are built upon the framework of the *artificial neural network* (ANN). ANNs are composed of a number of interconnected nodes (neurons), arranged in layers as shown in Figure 2. Neurons not contained in the input or output layers are referred to as hidden units. Every hidden unit stores a set of weights $W$ which are updated as the model is trained.

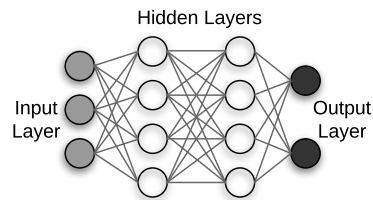

Fig. 2. Neural network with 1 input layer, 1 output layer, and 2 hidden layers.

ANN weights are optimized by minimizing a loss function such as the negative log likelihood, shown in Equation 1.

$$E(\theta, D) = -\sum_{i=0}^{D} \Big[ \log P(Y = y_i | x_i, \theta) \Big] + \lambda ||\theta||_p \quad (1)$$

The first term in Equation 1 minimizes the sum of the log loss across the entire training dataset $D$; the second term attempts to minimize the $p$-norm of the learned model parameters $\theta_i$ controlled by a tunable parameter $\lambda$. This second term is known as *regularization*, and is a technique used to prevent a model from overfitting and to increase its ability to generalize to new, unseen examples. The loss function is typically optimized using backpropagation, a mechanism for weight optimization that minimizes loss from the final layer backwards through the network [28].

Several open source tools exist for working with deep learning algorithms in a variety of programming languages,

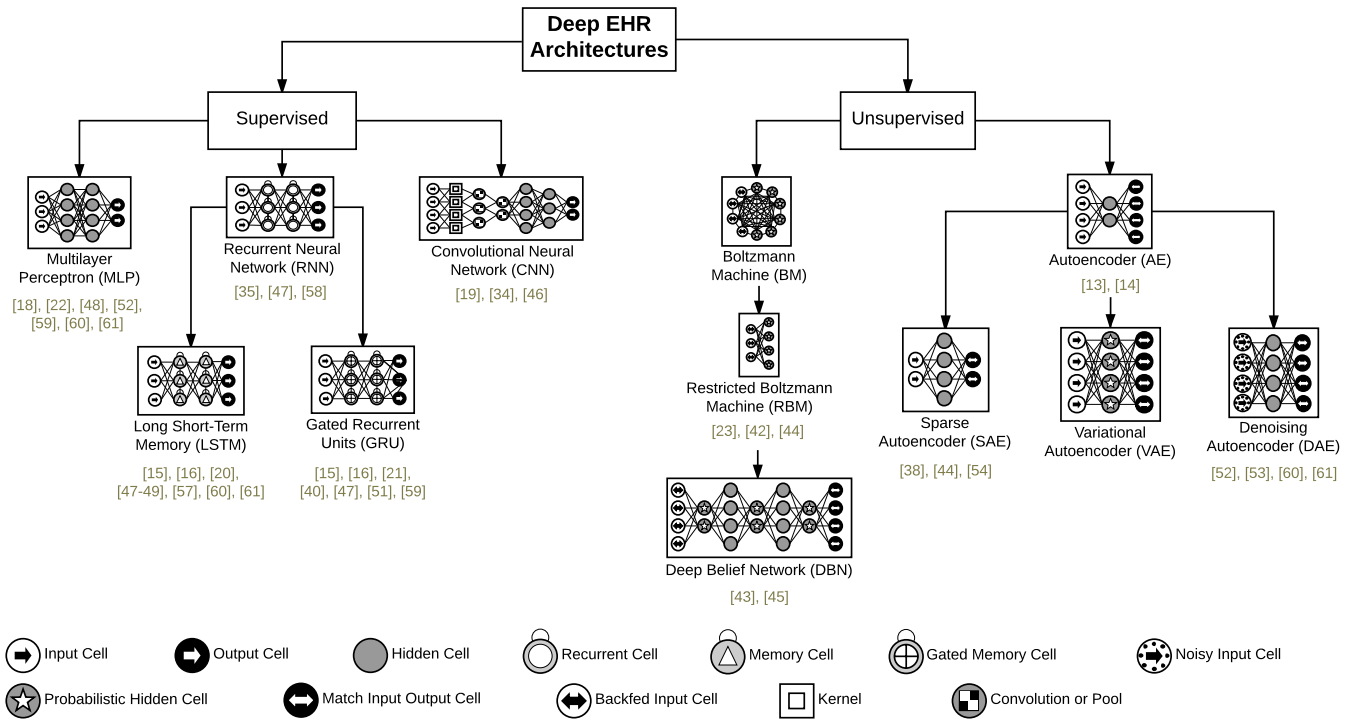

Fig. 3. The most common deep learning architectures for analyzing EHR data. Architectures differ in terms of their node types and the connection structure (e.g. fully connected versus locally connected). Below each model type is a list of selected references implementing the architecture for EHR applications. Icons based on the work of van Veen [30].

including TensorFlow[1], Theano[2], Keras[3], Torch[4], PyTorch[5], Caffe[6], CNTK[7], and Deeplearning4j[8].

In the remainder of this section, we review several common types of deep learning models used for deep EHR applications, all of which are based on the ANN's architecture and optimization strategy. We begin with supervised techniques (including multilayer perceptrons, convolutional neural networks, and recurrent neural networks) and conclude with unsupervised architectures (including autoencoders and restricted Boltzmann machines). A hierarchical view of these common deep learning architectures for analyzing EHR data, along with selected works in this survey which implement them, are shown in Figure 3.

*A. Multilayer perceptron (MLP)*

A multilayer perceptron is a type of ANN composed of multiple hidden layers, where every neuron in layer $i$ is fully connected to every other neuron in layer $i+1$. Typically, these networks are limited to a few hidden layers, and the data flows only in one direction, unlike recurrent or undirected models. Extending the notion of a single layer ANN, each hidden unit computes a weighted sum of the outputs from the previous layer, followed by a nonlinear activation $\sigma$ of the calculated sum as in Equation 2. Here, $d$ is the number of units in the previous layer, $x_j$ is the output from the previous layer's $j^{th}$ node, and $w_{ij}$ and $b_{ij}$ are the weight and bias terms associated with each $x_j$. Traditionally sigmoid or tanh are chosen as the nonlinear activation functions, but modern networks also use functions such as rectified linear units (ReLU) [28].

$$h_i = \sigma(\sum_{j=1}^{d} x_j w_{ij} + b_{ij}) \qquad (2)$$

After hidden layer weights are optimized during training, the network learns an association between input $x$ and output $y$. As more hidden layers are added, it is expected that the input data will be represented in an increasingly more abstract manner due to each hidden layer's nonlinear activations. While the MLP is one of simplest models, other architectures often incorporate fully connected neurons in their final layers.

*B. Convolutional neural networks (CNN)*

Convolutional neural networks (CNN) have become a very popular tool in recent years, especially in the image processing community. CNNs impose local connectivity on the raw data. For instance, rather than treating a 50x50 image as 2500 unrelated pixels, more meaningful features can be extracted by viewing the image as a collection of local pixel patches. Similarly, a one-dimensional time series can also be considered as a collection of local signal segments.

---
[1] https://www.tensorflow.org
[2] http://deeplearning.net/software/theano/
[3] https://keras.io
[4] http://torch.ch
[5] http://pytorch.org
[6] http://caffe.berkeleyvision.org
[7] https://www.microsoft.com/en-us/cognitive-toolkit/
[8] https://deeplearning4j.org

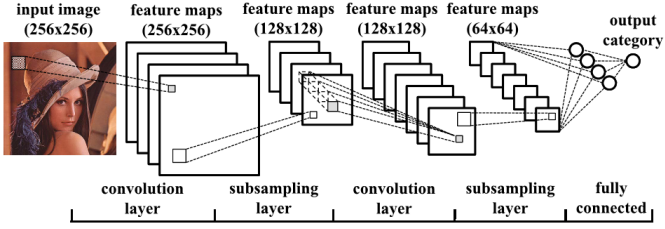

Fig. 4. Example of a convolutional neural network (CNN) for classifying images. This particular model includes two convolutional layers, each followed by a pooling/subsampling layer. The output from the second pooling layer is fed to a fully connected layer and a final output layer. [31]

The equation for one-dimensional convolution is shown in Equation 3, where $x$ is the input signal and $w$ is the weighting function or the convolutional *filter*.

$$C_{1d} = \sum_{a=-\infty}^{\infty} x(a)w(t-a) \qquad (3)$$

Similarly, two-dimensional convolution is shown in Equation 4, where $X$ is a 2-D grid (e.g., an image) and $K$ is a kernel. In this manner, a kernel or *filter* slides a matrix of weights across the entire input to extract the *feature maps*.

$$C_{2d} = \sum_m \sum_n X(m,n)K(i-m, j-n) \qquad (4)$$

CNNs involve *sparse* interactions as the filters are typically smaller than the input, resulting in relatively small number of parameters. Convolution also encourages *parameter sharing* since every filter is applied across the entire input.

In a CNN, a *convolution layer* is a number of convolutional filters described above, all receiving the same input from the previous layer, which ideally learn to extract different lower-level features. Following these convolutions, a *pooling* or *subsampling* layer is typically applied to aggregate the extracted features. An example CNN architecture with two convolutional layers, each followed by a pooling layer, is shown in Figure 4.

### C. Recurrent neural networks

While convolutional neural networks are a logical choice when the input data has a clear *spatial* structure (such as pixels in an image), recurrent neural networks (RNNs) are an appropriate choice when data is *sequentially* ordered (such as time series data or natural language). While one-dimensional sequences can be fed to a CNN, the resulting extracted features are shallow [28], in the sense that only closely localized relationships between a few neighbors are factored into the feature representations. RNNs are designed to deal with such long-range temporal dependencies.

RNNs operate by sequentially updating a hidden state $h_t$ based not only on the activation of the current input $x_t$ at time $t$, but also on the previous hidden state $h_{t-1}$, which in turn was updated from $x_{t-1}$, $h_{t-2}$, and so on (Figure 5). In this manner, the final hidden state after processing an entire sequence contains information from all its previous elements.

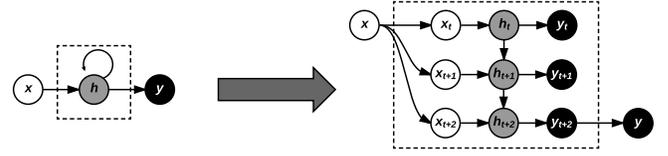

Fig. 5. Symbolic representation of a RNN (left) with equivalent expanded representation (right) for an example input sequence of length three, three hidden units, and a single output. Each input time step is combined with the current hidden state of the RNN, which itself depends on the previous hidden state, demonstrating the memory effect of RNNs.

Popular RNN variants include the long short-term memory (LSTM) and gated recurrent unit (GRU) models, both referred to as gated RNNs. Whereas standard RNNs are comprised of interconnected hidden units, each unit in a gated RNN is replaced by a special cell that contains an internal recurrence loop and a system of gates that controls the flow of information. Gated RNNs have shown benefits in modeling longer term sequential dependencies among other benefits [28].

### D. Autoencoders (AE)

One of the deep learning models exemplifying the notion of unsupervised representation learning is the autoencoder (AE). They were first popularized as an early tool to pretrain supervised deep learning models, especially when labeled data was scarce, but still retain usefulness for entirely unsupervised tasks such as phenotype discovery. Autoencoders are designed to *encode* the input into a lower dimensional space $z$. The encoded representation is then *decoded* by reconstructing an approximated representation $\tilde{x}$ of the input $x$. The encoding and reconstruction process for an autoencoder with a single hidden layer are respectively shown in Equations 5 and 6. $W$ and $W'$ are the respective encoding and decoding weights, and as the reconstruction error $\|x - \tilde{x}\|$ is minimized, the encoded representation $z$ is deemed more reliable.

$$z = \sigma(Wx + b) \qquad (5)$$

$$\tilde{x} = \sigma(W'z + b') \qquad (6)$$

Once an AE is trained, a single input is fed through the network, with the innermost hidden layer activations serving as the input's encoded representation. AEs serve to transform the input data into a format where only the most important derived dimensions are stored. In this manner, they are similar to standard dimensionality reduction techniques like principal component analysis (PCA) and singular value decomposition (SVD), but with a significant advantage for complex problems due to nonlinear transformations via each hidden layer's activation functions. Deep AE networks can be constructed and trained in a greedy fashion by a process called stacking (Figure 6). Many variants of AEs have been introduced, including denoising autoencoders (DAE) [32], sparse autoencoders (SAE), and variational autoencoders (VAE) [28].

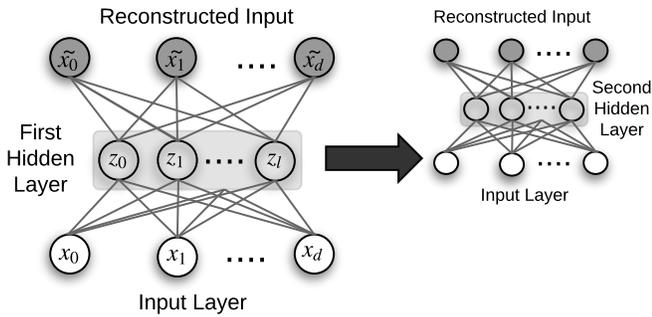

Fig. 6. Example of a stacked autoencoder with two independently-trained hidden layers. In the first layer, $\tilde{x}$ is the reconstruction of input $x$, and $z$ is lower dimensional representation (i.e., the encoding) of input $x$. Once the first hidden layer is trained, the embeddings $z$ are used as input to a second autoencoder, demonstrating how autoencoders can be stacked. [33]

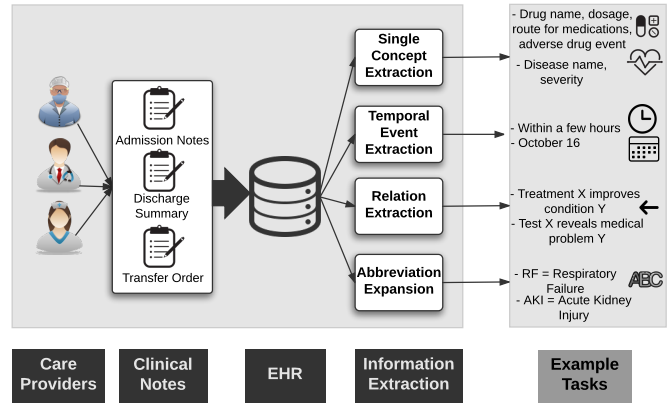

Fig. 7. EHR Information Extraction (IE) and example tasks.

### E. Restricted Boltzmann machine (RBM)

Another unsupervised deep learning architecture for learning input data representations is the restricted Boltzmann machine (RBM). The purpose of RBMs is similar to autoencoders, but RBMs instead take a stochastic perspective by estimating the probability distribution of the input data. In this way, RBMs are often viewed as generative models, trying to model the underlying process by which the data was generated.

The canonical RBM [28] is an energy-based model with binary visible units $\vec{v}$ and hidden units $\vec{h}$, with energy function specified in Equation 7.

$$E(v,h) = -b^T v - c^T h - W v^T h \qquad (7)$$

In a standard Boltzmann machine (BM), all units are fully connected, while in an RBM there are no connections between any two visible units or any two hidden units. Training an RBM is typically accomplished via stochastic optimization such as Gibbs sampling. It yields a final form of $h$, which can be viewed as the learned representation of the initial input data. RBMs can be hierarchically stacked to form a deep belief network (DBN) for supervised learning tasks.

## V. DEEP EHR LEARNING APPLICATIONS

In this section, we review the current state of the art in clinical applications resulting from recent advances in deep EHR learning. A summary of recent deep EHR learning projects and their target tasks is shown in Table III, where we propose task and subtask definitions based on a logical grouping of current research.

Many of the applications and results in the remainder of this section are based on private EHR datasets belonging to independent healthcare institutions, an issue we discuss further in Section VII. However, several studies included in this review make use of MIMIC (Medical Information Mart for Intensive Care), a freely-available critical care database[9], as well as public clinical note datasets from i2b2 (Informatics for Integrating Biology and the Bedside)[10].

[9]https://mimic.physionet.org
[10]https://www.i2b2.org/NLP/DataSets/

### A. EHR Information Extraction (IE)

In contrast to the structured portions of EHR data typically used for billing and administrative purposes, clinical notes are more nuanced and are primarily used by healthcare providers for detailed documentation. Each patient encounter is associated with several clinical notes, such as admission notes, discharge summaries, and transfer orders. Due to their unstructured nature, extracting information from clinical notes is very difficult. Historically these methods have required a large amount of manual feature engineering and ontology mapping, which is one reason why such techniques have seen limited adoption. As such, several recent studies have focused on extracting relevant clinical information from clinical notes using deep learning. The main subtasks include (1) single concept extraction, (2) temporal event extraction, (3) relation extraction, and (4) abbreviation expansion (Figure 7).

**(1) Single Concept Extraction**: The most fundamental task involving clinical free text is the extraction of structured medical concepts, such as diseases, treatments, or procedures. Several previous studies applied classical natural language processing (NLP) techniques to achieve this with varying levels of success, but there remains large room for improvement given the complexity of clinical notes. Jagannatha et al. [15], [16] treat the concept extraction problem as a sequence labeling task whose goal is to assign one of nine clinically relevant tags to each word in a clinical note. They divide tags into medication and disease categories, where each category contains relevant tags like drug name, dosage, route for medications, adverse drug event, indication, and severity of disease. They experiment with several deep architectures based on RNNs, including LSTMs and GRUs, bidirectional LSTMs (Bi-LSTM), and various combinations of LSTMs with traditional conditional random fields (CRF). In their experiments, they compare to baseline CRFs which had previously been considered the state-of-the-art technique for extracting clinical concepts from text. They found all variants of RNNs to outperform the CRF baselines by wide margins, especially in detecting more subtle attributes such as medication duration and frequency, and disease severity. Such nuanced information is highly important for clinical informatics and is not readily available from the billing-oriented clinical code structure. Other applications of deep learning to clinical concept extraction include named entity recognition (NER) in clinical text by Wu et al. [34], who apply pre-trained word embeddings on Chinese clinical text using a CNN, improving upon the CRF baselines.

TABLE III
SUMMARY OF EHR DEEP LEARNING TASKS.

| Task | Subtasks | Input Data | Models | References |
| --- | --- | --- | --- | --- |
| Information Extraction | (1) Single Concept Extraction<br>(2) Temporal Event Extraction<br>(3) Relation Extraction<br>(4) Abbreviation Expansion | Clinical Notes | LSTM, Bi-LSTM, GRU, CNN<br>RNN + Word Embedding<br>AE<br>Custom Word Embedding | [15], [16], [34]<br>[35]<br>[36]<br>[37] |
| Representation Learning | (1) Concept Representation<br>(2) Patient Representation | Medical Codes | RBM, Skip-gram, AE, LSTM<br>RBM, Skip-gram, GRU, CNN, AE | [23], [36]<br>[14], [18]–[23], [36], [38]–[40] |
| Outcome Prediction | (1) Static Prediction<br>(2) Temporal Prediction | Mixed | AE, LSTM, RBM, DBN<br>LSTM | [14], [18], [23], [41]–[43]<br>[19]–[21], [38], [44]–[48] |
| Phenotyping | (1) New Phenotype Discovery<br>(2) Improving Existing Definitions | Mixed | AE, LSTM, RBM, DBN<br>LSTM | [14], [40], [44], [49], [50]<br>[45], [51] |
| De-identification | Clinical text de-identification | Clinical Notes | Bi-LSTM, RNN + Word Embedding | [52], [53] |

**(2) Temporal Event Extraction**: This subtask tackles the more complex issue of assigning notions of time to each extracted EHR concept, such as *the last few months* or *October 16*. Fries [35] devised a framework to extract medical events and their corresponding times from clinical notes using a standard RNN initialized with word2vec [54] word vectors (explained in Section V-B) pre-trained on text from two large clinical corpora. Fries also utilizes Stanford's DeepDive application [55] for structured relationships and predictions. While not state of the art, it remained competitive in the SemEval 2016 shared task and required little manual engineering.

**(3) Relation Extraction**: While temporal event extraction associates clinical events with their corresponding time span or date, relation extraction deals with structured relationships between medical concepts in free text, including relations such as *treatment X improves/worsens/causes condition Y*, or *test X reveals medical problem Y*. Lv et al. [36] use standard text pre-processing methods and UMLS-based word-to-concept mappings in conjunction with sparse autoencoders to generate features for input to a CRF classifier, greatly outperforming the state of the art in EHR relation extraction.

**(4) Abbreviation Expansion**: There have been more than 197,000 unique medical abbreviations found in clinical text [37], which require expansion before mapping to structured concepts for extraction. Each abbreviation can have tens of possible explanations, thus making abbreviation expansion a challenging task. Liu et al. [37] tackle the problem by utilizing word embedding approaches. They create custom word embeddings by pre-training a word2vec model (Section V-B) on clinical text from intensive care units (ICU), Wikipedia, and medical articles, journals, and books. While word embedding models are not themselves *deep* models, they are a common prerequisite for NLP deep learning tasks. This embedding-based approach greatly outperformed baseline abbreviation expansion methodologies, scoring 82.3% accuracy compared with baselines in the 20-30% range. In particular, they found that combining all sources of background knowledge sources resulted in embeddings that yielded the greatest accuracy.

*Methods of evaluation for EHR information extraction*

Precision, recall, and F1 score were the primary classification metrics for the tasks involving single concept extraction [15], [16], [34], temporal event extraction [35], and clinical relation extraction [36]. The study on clinical abbreviation expansion [37] used accuracy as its evaluation method.

While some studies share similar tasks and evaluation metrics, results are not directly comparable due to proprietary datasets (discussed further in Section VII).

*B. EHR Representation Learning*

Modern EHR systems are populated by large numbers of discrete medical codes that reflect all aspects of patient encounters. Examples of codes corresponding to diagnoses, medications, laboratory tests, and procedures are shown in Table II. These codes were first implemented for internal administrative and billing tasks, but contain important information for secondary informatics.

Currently, handcrafted schema are used for mapping between structured medical concepts, where each concept is assigned a distinct code by its relevant ontology. These static hierarchical relationships fail to quantify the inherent similarities between concepts of different types and coding schemes. Recent deep learning approaches have been used to project discrete codes into vector space for more detailed analysis and more precise predictive tasks.

In this section, we first describe deep EHR methods for representing discrete medical codes (e.g. *I509*, corresponding to the ICD-10 code for heart failure) as real-valued vectors of arbitrary dimension. These projects are largely unsupervised and focus on natural relationships and clusters of codes in vector space. Since patients can be viewed as an ordered collection of medical event codes, in the following subsection we survey deep methods for representing *patients* using these codes. Patient representation frameworks are typically optimizing a supervised learning task (e.g. predicting patient mortality) by improving the representation of the input (e.g., the patients) to the deep learning network.

**(1) Concept Representation**: Several recent studies have applied deep unsupervised representation learning techniques to derive *EHR concept vectors* that capture the latent similarities and natural clusters between medical concepts. We refer to this area as EHR concept representation, and its primary objective is to derive vector representations from sparse medical codes such that similar concepts are nearby in the lower-dimensional vector space. Once such vectors are obtained, codes of heterogeneous source types (such as diagnoses and medications) can be clustered and qualitatively analyzed with techniques such as t-SNE [18], [19], [23], word-cloud visualizations of discriminative clinical codes [20], or code similarity heatmaps [40].

*Distributed Embedding*: Since clinical concepts are often recorded with a time stamp, a single encounter can be viewed as a sequence of discrete medical codes, similar to a sentence and its ordered list of words. Several researchers have applied NLP techniques for summarizing sparse medical codes into a fixed-size and compressed vector format. One such technique is known as skip-gram, a model popularized by Mikolov et al. in their word2vec implementation [54]. Word2Vec is an unsupervised ANN framework for obtaining vector representations of words given a large corpus where the representation of a word depends on the context, and this technique is often used as a pre-processing step with many text-based deep learning models. Similarly, E. Choi et al. [18], [22] and Y. Choi et al. [39] both use skip-gram in the context of clinical codes to derive distributed code embeddings. Skip-gram for clinical concepts relies on the sequential ordering of medical codes, and in the study of Y. Choi et al. [39], the issue of multiple clinical codes being assigned the same time stamp is handled by partitioning a patient's code sequence into smaller chunks, randomizing the order of events within each chunk, and treating each chunk as a separate sequence.

*Latent Encoding*: Aside from NLP-inspired methods, other common deep learning representation learning techniques have also been used for representing EHR concepts. Tran et al. formulate a modified restricted RBM which uses a structured training process to increase representation interpretation [23]. In a similar vein, Lv et al. use AEs to generate concept vectors from word-based concepts extracted from clinical free text [36]. They evaluated the strength of relationships between various medical concepts, and found that training linear models on representations obtained via AEs greatly outperformed traditional linear models alone, achieving state-of-the-art performance.

**(2) Patient Representation**: Several different deep learning methods for obtaining vector representations of patients have been proposed in the literature [20], [22], [23], [38], [39]. Most of the techniques are either inspired by NLP techniques such as distributed word representations [54], or use dimensionality reduction techniques such as autoencoders [13].

One such NLP-inspired approach is taken by Choi et al. [18], [22], [38] to derive distributed vector representations of patient *sentences*, i.e. ordered sequences of ICD-9, CPT, LOINC, and National Drug Codes (NDC), using both skip-gram and recurrent neural networks. Similarly, the Deepr framework uses a simple word embedding layer as input to a larger CNN architecture for predicting unplanned hospital readmission [19].

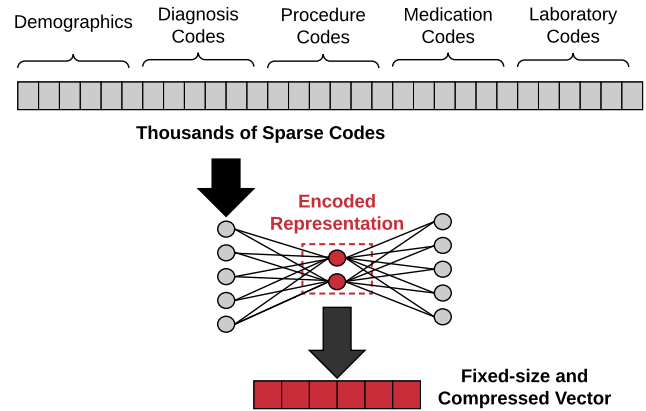

Fig. 8. Illustration of how autoencoders can be used to transform extremely sparse patient vectors into a more compact representation. Since medical codes are represented as binary categorical features, raw patient vectors can have dimensions in the thousands. Training an autoencoder on these vectors produces an encoding function to transform any given vector into it's distributed and dimensionality-reduced representation.

Miotto et al. directly generate patient vectors from raw clinical codes via stacked AEs, and show that their system achieves better generalized disease prediction performance as compared to using the raw patient features [14]. The raw features are vectorized via a three-layer AE network, with the final hidden layer's weights yielding the patient's corresponding representation. As a part of their framework for patient representation, they incorporated the clinical notes associated with each patient encounter into their representation framework using a form of traditional topic modeling known as latent Dirichlet allocation (LDA). An example of using autoencoders for patient representation is shown in Figure 8.

Choi et al. [18] derive patient vectors by first generating concept and encounter representations via skip-gram embedding, and then using the summed encounter vectors to represent an entire patient history to predict the onset of heart failure . Similarly, Pham et al in their DeepCare framework generate two separate vectors for a patient's temporal diagnosis and intervention codes, and obtain a patient representation via concatenation, showing that the resulting patient timeline vectors contain more predictive power than classifiers trained on the raw categorical features [20]. They employ modified LSTM cells for modeling time, admission methods, diagnoses, and interventions to account for complete illness history.

Aside from simple vector aggregation, it is also possible to directly model the underlying temporal aspects of patient timelines. Mehrabi et al. [40] use a stacked RBM trained on each patient's temporal diagnosis codes to produce patient representations over time. They pay special attention to temporal aspects of EHR data, constructing a diagnosis matrix for each patient with distinct diagnosis codes as rows, columns as binary variables indicating whether the patient was diagnosed with the code in a given time interval. Since rows of these matrices are clinical codes, the hidden layers of the RBM are the latent representations of the codes.

Finally, Choi et al.'s Doctor AI system utilizes sequences of (event, time) pairs occurring in each patient's timeline across multiple admissions as input to a GRU network [21]. At each time step, the weights of the hidden units are taken as the patient representation at that point in time, from which future patient statuses can be modeled and predicted.

*Methods of evaluation for EHR representation learning*

Many of the studies involving representation learning evaluate their representations based on auxiliary classification tasks, with the implicit assumption that improvements in prediction are attributed to a more robust representation of either clinical concepts or patients. Methods of evaluation are thus varied and task-dependent, including metrics such as AUC (heart failure onset prediction [18], [38], disease prediction [14], clinical risk group prediction [22]), precision@k (disease progression [20], disease tagging [14]), recall@k (medical code prediction [22], timed clinical event prediction [21]), accuracy (unplanned readmission prediction [19]), or precision, recall, and F1 score (relation extraction [36], unplanned readmission prediction [20], risk stratification [23]).

Some studies do not include any secondary classification tasks, and focus on evaluating the learned representations directly. As there is no agreed-upon metric for such tasks, evaluation methods are again varied. Tran et al. use the notion of concept coherence, originally seen in topic modeling [23]. Choi et al. introduce two custom metrics referred to as *medical concept similarity measure (MCSM)* and *medical relatedness measure (MRM)* to quantitatively evaluate clusters of clinical codes [39].

While these are two distinct methods for quantitatively evaluating clinical representations, research from both types share a common component of qualitative analysis. This typically involves subjectively evaluating similarity between representations of either concepts or patients in the embedded vector space, visualized with techniques such as t-SNE [19], [23] or plotted via heatmap clusters [40].

While some studies share similar tasks and evaluation metrics, results are not directly comparable due to proprietary datasets (discussed further in Section VII).

*C. Outcome Prediction*

The ultimate goal of many Deep EHR systems is to predict patient outcomes. We identify two different types of outcome prediction: (1) *static* or one-time prediction (e.g. heart failure prediction using data from a single encounter), and (2) *temporal* outcome prediction (e.g. heart failure prediction within the next 6 months, or disease onset prediction using historical data from sequential encounters). Many of these prediction frameworks make use of unsupervised data modeling, such as clinical concept representation (Section V-B). In many cases, the main contribution is the deep representation learning itself, with an increase in performance using linear models being used for assessing the quality of the derived representations.

TABLE IV
OUTCOME PREDICTION TASKS IN DEEP EHR PROJECTS.

| Outcome Type | Outcome | Model |
| --- | --- | --- |
| Static | Heart Failure | MLP [18] |
| | Hypertension | CNN [41] |
| | Infections | RBM [42] |
| | Osteoporosis | DBN [43] |
| | Suicide risk stratification | RBM [23] |
| Temporal | Cardiovascular, Pulmonary | CNN [44] |
| | Diabetes, Mental Health | LSTM [20] |
| | Re-admission | TCNN [19] |
| | Heart Failure | GRU [21], [38] |
| | Renal | RNN [47] |
| | Postoperative Outcomes | LSTM [46] |
| | Multi-outcome (78 ICD codes) | AE [14] |
| | Multi-outcome (128 ICD codes) | LSTM [45] |

**(1) Static Outcome Prediction**: The simplest class of outcome prediction applications is the prediction of a certain outcome without considering temporal constraints. For example, Choi et al. use distributed representations and several ANN and linear models to predict heart failure [18]. They found the best model to be a standard MLP trained with the embedded patient vectors, outperforming all variants using the raw categorical codes.

Tran et. al [23] derive patient vectors with their modified RBM architecture, then train a logistic regression classifier for suicide risk stratification. They experimented with using the full EHR data vs. only using diagnosis codes, and found that the classifier using the complete EHR data with the eNRBM architecture for concept embeddings performed best. Similarly, DeepPatient generated patient vectors with a 3-layer autoencoder, then used these vectors with logistic regression classifiers to predict a wide variety of ICD9-based disease diagnoses within a prediction window [14]. Their framework showed improvements over raw features, with superior precision@k metrics for all values of k. In a conceptually similar fashion, Liang et al. [41] also generated patient vectors for use with linear classifiers, but opted for layer-wise training of a Deep Belief Network (DBN) followed by a support vector machine (SVM) for classifying general disease diagnoses.

Since ideally clinical notes associated with a patient encounter contain rich information about the entirety of the admission, many studies have examined outcome prediction from the text alone. Jacobson et al. [42] compared deep unsupervised representation of clinical notes for predicting healthcare-associated infections (HAI), utilizing stacked sparse AEs and stacked RBMs along with a word2vec-based embedding approach. They found that a stacked RBM with term frequency-inverse document frequency (tf-idf) pre-processing yielded the best average F1 score, and that applying word2vec pre-training worked better with the AEs than the RBMs.

Finally, Li et al. [43] used a two-layer DBN for identifying osteoporosis. Their framework used a discriminative learning stage where top risk factors were identified based on DBN reconstruction errors, and found the model using all identified risk factors resulted in the best performance over baselines.

**(2) Temporal Outcome Prediction**: Other studies have trained deep learning architectures with the primary purpose of *temporal* outcome prediction, either to predict the outcome or onset within a certain time interval or to make a prediction based on time series data. Cheng et al. trained a CNN on temporal matrices of medical codes per patient for predicting the onset of both congestive heart failure (CHF) and chronic obstructive pulmonary disease (COPD) [44]. They experimented with several temporal CNN-specific techniques such as slow, early, and late fusion, and found that the CNN with slow fusion outperformed other CNN variants and linear models for both prediction tasks.

Lipton et al. used LSTM networks for predicting one of 128 diagnoses, using target replication at each time step along with auxiliary targets for less-common diagnostic labels as a form of regularization [45]. Among the deep architectures, they found the best diagnostic performance occurred with a two-layer LSTM of 128 memory cells each. The best overall performance was achieved with an ensemble framework with their top LSTM in conjunction with a standard three-layer MLP using more traditional handcrafted features.

Choi et al.'s Doctor AI framework was constructed to model how physicians behave by predicting future disease diagnosis along with corresponding timed medication interventions [21]. They trained a GRU network on patients' observed (clinical event, time) pairs, with the goal of predicting the next coded event, along with its time, and any future diagnoses. They found that their system performed differential diagnosis with similar accuracy to physicians, achieving up to 79% recall@30 and 64% recall@10. Interestingly, they also found their system performed similarly well using a different institution's coding system, and found that performance on the publicly available MIMIC dataset [56] was increased by pre-training their models on their own private data. They then expanded their work [38] by training a GRU network on sequences of clinical event vectors derived from the same skip-gram procedure, and found superior performance over baselines for predicting the onset of heart disease during various prediction windows.

Pham et al.'s DeepCare framework also derives clinical concept vectors via a skip-gram embedding approach, but creates two separate vectors per patient admission: one for diagnosis codes, and another for intervention codes [20]. The concatenation of these vectors is passed into an LSTM network for predicting the next diagnosis and next intervention for both diabetes and mental health cohorts. They model disease progression by examining precision@k metrics for all prediction tasks. They also predict future readmission based on these past diagnoses and interventions. For all tasks, they found the deep approaches resulted in the best performance.

Nickerson et al. [46] forecast postoperative responses including postoperative urinary retention (POUR) and temporal patterns of postoperative pain using MLP and LSTM networks to suggest more effective postoperative pain management.

Nguyen et al. [19]'s Deepr system uses a CNN for predicting unplanned readmission following discharge. Similar to several other methods, Deepr operates with discrete clinical event codes. They examine the clinical motifs arising from the convolutional filters as a form of interpretability, and found their methods to be superior to the bag-of-codes and logistic regression baseline models. Interestingly, they found that large time gaps in the input sequences do not affect the accuracy of their system, even though they did not specifically pre-process their data to account for them.

Esteban et al. [47] used deep models for predicting the onset of complications relating to kidney transplantation. They combined static and dynamic features as input to various types of RNNs, binning the continuous laboratory measurements as low, normal, or high. They found a GRU-based network, in conjunction with static patient data, outperformed other deep variants as well as linear baseline models. They also found that using embeddings of static features resulted in better performance for tasks where long-term dependencies were not as important, but dynamic embeddings were more useful for inputs with significant time dependencies.

Finally, Che. et al [48] develop a variation of the recurrent GRU cell (GRU-D) which attempts to better handle missing values in clinical time series, citing a frequent correlation in literature between missing values and outcome. Their GRU-D networks show improved AUC on two real-world ICD-9 classification and mortality prediction tasks.

*Methods of evaluation for EHR outcome prediction*

While outcome prediction tasks are widely varied, most of the methods of evaluation for outcome prediction using deep learning techniques make use of standard classification metrics such as AUC (heart failure prediction [18], [38], diagnosis classification [14], [44], [45], [51], bone disease risk factor identification [43], clinical event prediction [47]), accuracy (predicting analgesic response [46], unplanned readmission prediction [19]), and precision, recall, and F1 score (risk stratification [23], hypertension prediction [41], diagnosis prediction [45]). For tasks involving temporal prediction, we also see metrics such as precision@k and recall@k (temporal diagnosis prediction [45], disease progression modeling [20], timed clinical event prediction [21]).

While some studies share similar tasks and evaluation metrics, results are not directly comparable due to proprietary datasets (discussed further in Section VII).

*D. Computational Phenotyping*

As the amount and availability of detailed clinical health records has exploded in recent years, there is a large opportunity for revisiting and refining broad illness and diagnosis definitions and boundaries. Whereas diseases are traditionally defined by a set of manual clinical descriptions, computational phenotyping seeks to derive richer, *data-driven* descriptions of illnesses [51]. By using machine learning and data mining techniques, it is possible to discover natural clusters of clinical descriptors that lend themselves to a more fine-grained disease description. Detailed phenotyping is a large step towards the eventual goal of personalized and precision healthcare.

Computational phenotyping can be seen as an archetypal clinical application of deep learning principles, which is grounded in the philosophy of *letting the data speak for itself* by discovering latent relationships and hierarchical concepts from the raw data, without any human supervision or prior bias. With the availability of huge amounts of clinical data, many recent studies have employed deep learning techniques for computational phenotyping.

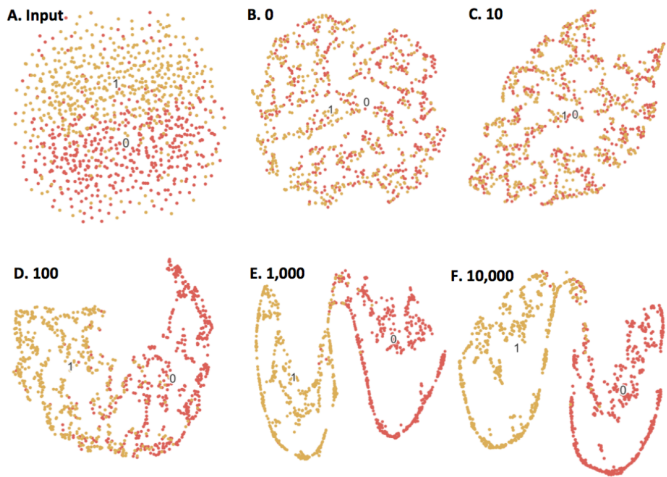

Fig. 9. Beaulieu-Jones and Greene's [49] autoencoder-based phenotype stratification for case (1) vs. control (0) diagnoses, illustrated with t-SNE. (A) shows clustering based on raw clinical descriptors, where there is little separable structure. (B-F) show the resulting clusters following 0-10,000 training epochs of the single-layer autoencoder. As the autoencoder is trained, there are clear boundaries between the two labels, suggesting the unsupervised autoencoder discovers latent structure in the raw data without any human input.

Computational phenotyping research is composed of two primary applications: (1) discovering and stratifying new subtypes, and (2) discovering specific phenotypes for improving classification under existing disease boundaries and definitions. Both areas seek to discover new data-driven phenotypes; the former is a largely unsupervised task that is difficult to quantitatively evaluate, where the latter is inherently tied to a supervised learning task whose results can be easily validated.

**(1) New Phenotype Discovery**: As phenotyping is a largely unsupervised task, several recent studies have utilized AEs for discovering phenotypes from raw data, since enforcing a lower-dimensional data representation encourages discovery of latent structure. In perhaps the most straightforward application, Beaulieu-Jones and Greene employed a single-layer DAE for encoding patient records comprised of various binary clinical descriptors [49]. Figure 9 shows t-SNE visualizations (Section VI) for their phenotype-based stratification for a simulated diagnosis. They found that when paired with a random forest classifier, the AE representation had competitive accuracy with SVMs and decision trees while using a much smaller feature space, suggesting a latent structure in the input features. They also found that DAEs were much more robust to missing data, which is often an issue in practice.

A drawback of Beaulieu-Jones and Greene's work [49] is that the 20,000-patient dataset was synthetically constructed under their own simulation framework. Miotto et. al [14] devised a similar but more complex approach to patient representation based on AEs, using 704,587 real patient records from the Mount Sinai data warehouse. Whereas the clinical descriptors in Beaulieu-Jones and Greene's study [49] were entirely simulated, Miotto et al.'s DeepPatient framework [14] uses a combination of ICD-9 diagnoses, medications, procedures, lab tests, and conceptual topics from clinical free text obtained from latent Dirichlet allocation (LDA) as input to their AE framework. Compared to Beaulieu-Jones's single hidden layer, DeepPatient adds two more hidden layers for discovering additional complexity. Using a simple logistic regression classifier, the AE patient representations were compared with the raw features as well as those obtained from other dimensionality reduction techniques such as principal component analysis (PCA) and independent component analysis (ICA), where DeepPatient showed improvements in ICD9-based diagnosis prediction accuracy over 30, 60, 90, and 180-day prediction windows.

Cheng et al. [44] used a CNN model which yielded superior phenotypes and classification performance over baselines, with their slow fusion variants performing the best. They represent patient data as a temporal matrix with time on one axis and events on the other. They build a four-layer CNN model for extracting phenotypes and perform prediction. The first layer is composed of those EHR matrices. The second layer is a one-side convolution layer that can extract phenotypes from the first layer. The third layer is a max pooling layer introducing sparsity on the detected phenotypes, so that only those significant phenotypes will remain. The fourth layer is a fully connected softmax prediction layer.

Similar to the work of Cheng et al. [44], Mehrabi et al. also construct patient matrices from discrete codes using an RBM, and take its fist hidden layer as the embedded patient representation [40]. They found natural clusters of related codes and examined how the corresponding phenotypes change over time.

In the previously mentioned phenotyping applications, patient data came in the form of a set of discrete codes, from which distributed embeddings were created using deep feature representation. However, other phenotyping studies use continuous time-series data rather than static codes represented as one-hot vectors. Lasko et al. examined the problem of phenotyping continuous-time uric acid measurements, for distinguishing between gout and acute leukemia diagnoses [50]. They applied Gaussian processes and time warping to pre-process the time series, followed by a two-layer stacked AE for extracting specific phenotypes from 30-day patches of uric acid measurements. They found the first layer of the AE learned functional element detectors and basic trajectory patterns, and the overall embeddings were found to generate identifiable clusters in t-SNE representations, suggesting the presence of subtypes that should be explored in future work.

**(2) Improving Existing Definitions**: This class of algorithms typically try to improve current phenotypes by using a supervised learning approach. For example, Lipton et al. [45] utilize multivariate time series consisting of 13 variables from the ICU to predict phenotypes. They frame the problem of phenotyping as a multi-label classification problem, as phenotypes are traditionally composed of several binary indicators. They used an LSTM network with target replication at each time step for predicting a multi-label output from the 100 most frequent diagnoses in their data, and introduced auxiliary target objectives for the less-common diagnostic labels as a form of regularization. As compared with logistic regression and MLP baselines, their approach was superior, and they found that an ensemble of LSTM + MLP yielded the best performance.

Finally, Che et al. [51] also develop computational phenotyping techniques for working with multivariate clinical time series data, and as in the study of Lipton et al. [45], treat the phenotyping task as a multi-label classification problem. They use a standard MLP architecture pre-trained with DAE, but introduce a prior-based Laplacian regularization process on the final sigmoid layer that is based on structured medical ontologies, citing that using prior knowledge is especially useful for rare diseases with few training cases. They also develop a novel incremental training procedure that iteratively adds neurons to hidden layers. They examine the derived phenotype features using a standard maximum activation analysis of the hidden units, identifying several key patterns in time series most notably for both circulatory disease and septic shock.

*Methods of evaluation for computational phenotyping*

Similar to evaluating concept and patient representations, the methods for evaluating phenotype classifications share quantitative and qualitative components. Several deep phenotyping studies employ the use of an external prediction task to implicitly evaluate the strength of the phenotypes, and make use of classification metrics such as AUC, accuracy, precision, recall, and F1 score (simulated disease prediction [49], chronic disease onset prediction [44], gout vs. leukemia prediction [50], diagnosis prediction [45], clinical event prediction [51]). Qualitative methods are also similar, involving subjectively identifying patient clusters using t-SNE [49], [50] or identifying heatmap clusters [40].

While some studies share similar tasks and evaluation metrics, results are not directly comparable due to proprietary datasets (discussed further in Section VII).

*E. Clinical Data De-identification*

Clinical notes typically include explicit personal health information (PHI), which makes it difficult to publicly release many useful clinical datasets [57]. According to the guidelines of the Health Information Portability and Accountability Act (HIPAA), all the clinical notes released must be free of sensitive information such as names of patients and their proxies, identification numbers, hospital names and locations, geographic locations and dates [58]. Dernoncourt et al. [52] created a system for the automatic de-identification of clinical text, which replaces a traditionally laborious manual de-identification process for sharing restricted data. Their framework consists of a bidirectional LSTM network (Bi-LSTM) and both character and word-level embeddings. The authors found their method to be state of the art, with an ensemble approach with conditional random fields also faring well.

In a similar task, Shweta et al. [53] explore various RNN architectures and word embedding techniques for identifying potentially identifiable named entities in clinical text. The authors demonstrate that all RNN variants outperform traditional CRF baselines on a clinical shared task dataset, with location-based PHI proving the most difficult to accurately detect.

TABLE V
INTERPRETABILITY TECHNIQUES FOR DEEP EHR SYSTEMS

| Type | Methods |
|---|---|
| (1) Maximum activation | Convolutional filter response [19]<br>Output activation maximization [22]<br>Dense top-layer weight maximization [44] |
| (2) Constraints | Non-negativity [23]<br>Non-negative matrix factorization [22]<br>Sparsity [50]<br>Ontology smoothing [23]<br>Regularization [23] |
| (3) Qualitative clustering | Principal component analysis [49]<br>t-SNE [19] |
| (4) Mimic learning | Interpretable mimic learning [59]–[61] |

*Methods of evaluation for clinical data de-identification*

These tasks involve predicting a PHI category for each word in a clinical note, and evaluate their frameworks based on precision, recall, and F1 score. The datasets used in these studies are both open source, using clinical notes from MIMIC [52] and i2b2: Informatics for Integrating Biology and the Bedside [52], [53].

VI. INTERPRETABILITY

While deep learning techniques have gained notoriety for producing state-of-the-art performance on a variety of tasks, one of its main criticisms is that the resulting models are difficult to naturally interpret. In this regard, many deep learning frameworks are often referred to as "black boxes", where only the input and output predictions convey meaning to a human observer. The main culprit for this lack of model transparency is precisely what makes deep learning so effective: the layers of nonlinear data transformations that uncover hidden factors of entanglement in the input. This conundrum presents a tradeoff between performance and openness.

In the clinical domain, model transparency is of utmost importance, given that predictions might be used to affect real-world medical decision-making and patient treatments. This is one reason why interpretable linear models such as logistic regression still dominate applied clinical informatics. In fact, many studies in this paper have explicitly mentioned the lack of interpretability as a main limitation [14], [17], [21], [45], [62]. In this section, we briefly review attempts to make clinical deep learning more interpretable.

**(1) Maximum Activation**: A popular tactic in the image processing community is to examine the types of inputs that result in the maximum activation of each of a model's hidden units. This represents an attempt to examine what exactly the model has learned, and can be used to assign importance to the raw input features. This approach has been adopted by several studies included in our overview [19], [22], [44], [51].

**(2) Constraints**: Others have imposed training constraints specifically aimed at increasing the interpretability of deep

models. Choi et al.'s Med2Vec framework [22] for learning concept and patient visit representations uses a non-negativity constraint enforced upon the learned code representations. The authors take the $k$ largest values of each column of the resulting code weight matrix as a distinct disease group that is interpretable upon qualitative inspection. They also perform the same process on the resulting visit embedding matrix for analyzing the types of visits each neuron learns to identify.

Similarly, Tran et al.'s eNRBM architecture [23] also enforces non-negativity in the weights of the RBM. The authors claim that the resulting weight sparsity is an indication of which types of inputs activate the sparse neurons and can be used as a technique to estimate the intrinsic dimensionality of the data. They also develop a novel regularization framework for promoting structural smoothness based on the structure of medical ontologies, by encoding the hierarchies into a feature graph with edge weights based on the ontology distances. Both of these constraints are added to the overall objective function for generating vector representations of medical concepts.

In Lasko et al.'s phenotype discovery framework for time series data [50], their regularization and sparsity constraints on the AE yielded continuous features on the first layer that were interpretable as functional element detectors such as uphill or downhill signal ramps (Figure 10), another example of how sparsity improves interpretability of learned model weights.

**(3) Qualitative Clustering**: In the case of EHR concept representation and phenotype studies, some studies point to a more indirect notion of interpretability by examining natural clusters of the resulting vectorized representations. This is most commonly performed using a visualization technique known as t-Distributed Stochastic Neighbor Embedding (t-SNE), a method for plotting pairwise similarities between high-dimensional data points in two dimensions [63]. Beaulieu-Jones et al. [49] first perform principal component analysis (PCA) on the hidden weights of their autoencoder, followed by t-SNE to examine the clusters and separability of case vs. control examples (Figure 9), which show a clear phenotype distinction. The authors leave a detailed examination of the components of the phenotypes for future work.

In a similar fashion, Nguyen et al. [19] project distributed representations of both clinical event and patient vectors into two dimensions via t-SNE, allowing for a qualitative comparison of similar diagnoses and patient subgroups. Tran et al. [23] perform a similar visualization for medical objects.

**(4) Mimic Learning**: Finally, Che et al. [59]–[61] tackle the issue of deep model transparency in their Interpretable Mimic Learning frameworks. They first train a deep neural network on raw patient data with associated class labels, which produces a vector of class probabilities for each sample. They train an additional gradient boosting tree (GBT) model on the raw patient data, but instead use the deep network's probability prediction as the target label. Since GBTs are interpretable linear models, they are able to assign feature importance to the raw input features while harnessing the power of deep networks. The mimic learning method is shown to have similar or better performance than both baseline linear and deep models for select phenotyping and mortality prediction tasks, while retaining the desired feature transparency.

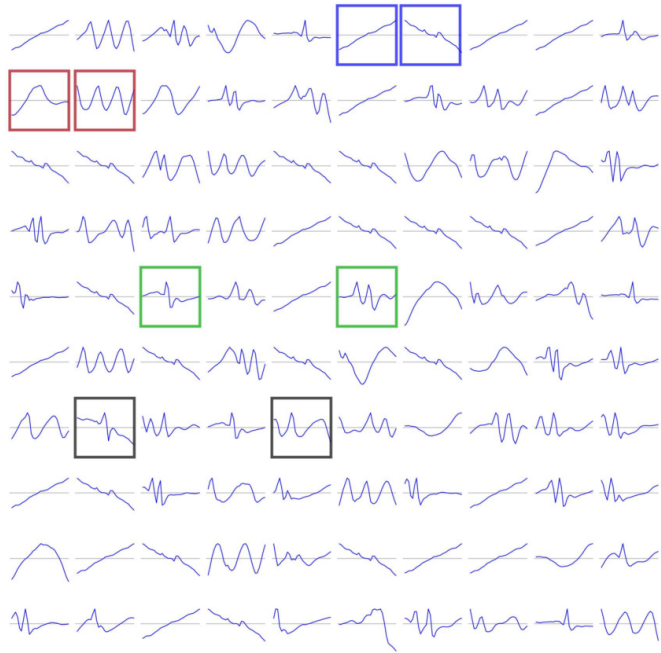

Fig. 10. Example of the positive effect of sparsity constraints on model interpretability. Shown are the first hidden layer weights from Lasko et al.'s [50] autoencoder framework for phenotyping uric acid sequences, which in effect form functional element detectors.

## VII. DISCUSSION AND FUTURE DIRECTION

In this paper, we have provided a brief overview of current deep learning research as it pertains to EHR analysis. This is an emerging area as evidenced by the fact that most of the papers we have surveyed were published in the past two years.

Tracing back the deep learning-based advances in image and natural language processing, we see a clear chronological similarity to the progression of current EHR-driven deep learning research. Namely, a majority of studies in this survey are concerned with the idea of representation learning, i.e., how best to represent the vast amounts of raw patient data that has suddenly become available in the past decade. Fundamental image processing research is concerned with increasingly complex and hierarchical representations of images composed of individual pixels. Likewise, NLP is focused on word, sentence, and document-level representations of language composed of individual words or characters. In a similar fashion, we are seeing the exploration of various schemes of representing patient health data from individual medical codes, demographics, and vital signs. The parallels are strong, and these recent studies represent a critical launching off point for future deep clinical research.

**Data Heterogeneity**: While we can draw clear similarities to other domains, what makes EHR representation learning (and EHR deep learning in general) so unique, and perhaps so difficult, is the variety of forms in which data is available. As we have seen, EHR data is quite heterogeneous, in clear contrast to the homogeneity of raw inputs to image processing (pixels) or NLP (characters) frameworks. EHR patient data can arise not only in the form of free text from clinical notes and radiological reports, but also as discrete billing-centric medical codes, patient demographic information, continuous time-series of vital signs and other laboratory measurements, medication dosages of varying potency, and more.

As the field begins to take shape, what we have seen so far is primarily a divide-and-conquer approach to representation learning for dealing with this mixed-type data. As perhaps the most logical starting point, several studies examined only the set of discrete medical codes associated with each patient encounter, a tractable way to process EHR events encompassing a variety of source data types like diagnoses, procedures, laboratory tests, medications, and vital signs. Current research in this area exhibits a strong parallel to NLP, in which clinical codes are viewed as "words" and patients or encounters as the "sentences". In some cases, sequences of patient encounters can also be seen as the "documents".

Several studies focused on deriving vector-based representation of clinical concepts to reduce the dimensionality of the code space and reveal latent relationships between similar types of discrete codes. Following the NLP analogy, these techniques resemble word embedding methods such as word2vec [54]. Deep EHR methods for code representation include the NLP-inspired skip-gram technique for predicting heart failure [18] and standalone code clustering [39], a CNN model to predict unplanned readmissions [19], an RBM-based framework for stratifying suicide risk [23], DBMs for diagnosis clustering [40], and a technique based on LSTMs for modeling disease progression and predicting future risk for diabetes and mental health [20]. In many of these studies, code representations were evaluated based on an auxiliary prediction task, in which the aforementioned methods outperformed traditional baseline approaches. Each of these studies also included qualitative analysis of code clusters using either t-SNE [18], [19], [23], word-cloud visualizations of discriminative clinical codes [20], or code similarity heatmaps [40].

Moving up a layer in the hierarchy, other studies focused on representing patients using combinations or sequences of their reported clinical codes. Techniques here range from autoencoders for diagnosis prediction [14], CNNs for unplanned readmission prediction [19], GRU networks for predicting future clinical events [21], LSTMs for future risk prediction [20], and multi-layer recurrent neural networks for disease prediction [22]. Similar to code representation, these studies typically involve an auxiliary classification task to evaluate the robustness of patient representations. For unsupervised techniques like autoencoders, patient representations are learned independently of the prediction task; with supervised models like CNNs or RNN variants, the representations are typically learned jointly with the predictive model.

While code-based representations of clinical concepts and patient encounters are a tractable first step towards working with heterogeneous EHR data, they ignore many important real-valued measurements associated with items such as laboratory tests, intravenous medication infusions, vital signs, and more. In the future, we expect more research to focus on processing these diverse sets of data directly, rather than relying on codes from controlled vocabularies that are primarily designed for billing purposes.

**Irregular Measures**: Aside from code-based representations, other studies have approached EHR data from a signal processing standpoint, focusing on the wealth of continuous time series data available in the form of vital signs and other timestamped measurements like laboratory test results. Such research includes using LSTMs with vital signs for predicting diagnoses [45], autoencoders with uric acid measurements for distinguishing between gout and leukemia [50], and MLPs for predicting in-hospital mortality from multivariate ICU time series [51]. The primary concern with this type of framework is the irregularity of scale - some signals are measured on a sub-hourly basis while others are on a monthly or yearly time scale. Currently time-based pre-processing is important in these types of studies. Even given the narrow focus on individual temporal variables, we see deep learning's impact on the ability to identify distinct patterns for applications like phenotype discovery from vital signs [50]. We expect future deep learning research involving clinical time series to include more robust mechanisms for handling irregularly sampled measurements without any handcrafted preprocessing procedures.

**Clinical Text**: Perhaps the most untapped resource for future deep clinical methods is the often staggering amount of free text associated with each patient encounter, appearing in the form of admission and discharge summaries, clinician notes, transfer requests, and more. This text contains a wealth of information about each patient, but extracting it is a difficult problem compounded by its often complete lack of structure. As an example, the same type of note can appear very differently depending on its author, due to various shorthand abbreviations, ordering preferences, and writing style. Properly utilizing these notes is an open problem that will necessarily draw from fundamental NLP techniques as well as those specific to the clinical domain.

The first step to dealing with clinical free text is the extraction of structure, and indeed we have seen several deep learning-based advances in the realm of information extraction from clinical notes, with techniques ranging from LSTM-CRFs [15] and Bi-RNNs [16] for extracting clinical concepts and events, DNNs for clinical named entity recognition [34], autoencoders for medical concept relation extraction [36], word embeddings for clinical abbreviation expansion [37], and RNNs for extracting temporal expressions [35]. Additionally, sparse deep learning has been used for disease inference from natural language health questions [17]. Given the amount of knowledge contained in clinical text, we feel there is still great opportunity for further text-based clinical informatics research.

**Unified Representation**: It appears that the next logical step for deep EHR research is the development of frameworks that utilize *all* types of patient data, not sets of homogenous data types considered in isolation. Given the results of the individual systems discussed in this paper, a truly unified patient representation appears to be one of the holy grails of clinical deep learning research. While deep learning research based on mixed data types is still ongoing, this type of universal representation learning will have huge benefits for patient and disease modeling, patient trajectory prediction, intervention recommendation, and more.

**Patient De-identification**: Since Deep EHR frameworks

require large amounts of data to perform effectively, the transfer of information between hospitals and institutions is a necessity for future model improvements. However, strict privacy policies prevent disclosure of sensitive patient data. At current time, only a few studies have explored deep learning techniques for automatic patient de-identification, including methods such as RNNs [53] and bidirectional LSTMs with character-enhanced embeddings [52], both of which show improvements over traditional de-identification models based on lexical features. Given the potential for facilitating cross-institutional knowledge sharing, we expect deep patient de-identification to be another large area of future research.

**Benchmarks**: Another key Deep EHR issue that must be addressed is the lack of transparency and reproducibility of reported results. Most of the studies in this paper use their institution's own private dataset, and given the sensitive nature of patient health data, they are understandably hesitant about sharing their data or making it public. However, a drawback to these policies is the lack of universally agreed-upon reference benchmarks for incremental algorithm improvements. Many studies claim state-of-the-art results, but few can be verified by external parties. This is a barrier for future model development and one cause of the slow pace of advancement.

**Interpretability**: Finally, we note that while in many cases predictive models are improved by using deep learning methodologies, the human interpretability of such models remains an elusive goal. As discussed in Section VI, model transparency is of utmost importance to many clinical applications, and as such, we expect this to be a large focus of clinical deep learning research moving forward. Since correct clinical decision-making can be the difference between life and death, many practitioners must be able to understand and trust the predictions and recommendations made by deep learning systems. While some researchers downplay the importance of interpretability in favor of significant improvements in model performance, we feel advances in deep learning transparency will hasten the widespread adoption of such methods in clinical practice. Context also plays a role, where life-or-death decisions from systems with only marginal improvements in accuracy over a human practitioner may warrant greater transparency than systems with near-perfect accuracy, or those with lower stakes. As such, we expect deep EHR interpretability to remain an ongoing area of future research.